\documentclass{article}
\usepackage[utf8]{inputenc} 

\usepackage{graphicx}
\usepackage{longtable}
\usepackage{hyperref}

\usepackage{arxiv}
\usepackage{amsmath}
\usepackage{algorithm}
\usepackage{algpseudocode}
\usepackage{algorithmicx}
\usepackage[utf8]{inputenc} 
\usepackage[T1]{fontenc}    
\usepackage{hyperref}       
\usepackage{url}            
\usepackage{booktabs}       
\usepackage{amsfonts}       
\usepackage{nicefrac}       
\usepackage{microtype}      
\usepackage{lipsum}
\usepackage{amsmath}
\usepackage{graphicx}
\graphicspath{ {./images/} }
\usepackage{tabularx}
\usepackage{subcaption}
\title{Quantum Down Sampling Filter for Variational Auto-encoder 
}
\author{
 Farina Riaz \\
  CSIRO Data61\\
  Sydney \\
  \texttt{farina.riaz@csiro.au} \\
   \And
 Fakhar Zaman\\
  CSIRO Manufacturing\\
 Clayton Victoria \\
  \texttt{fakhar.mit@gmail.com} \\
  \And
 Hajime Suzuki \\
  CSIRO Data61\\
  Sydney\\
  \texttt{hajime.suzuki@csiro.au} \\
\And
Sharif Abuadbba \\
 CSIRO Data61\\
 Sydney\\
 \texttt{sharif.abuadbba@csiro.au} \\
 \And
David Nguyen \\
 CSIRO Data61\\
 Sydney\\
 \texttt{david.nguyen@csiro.au} \\  
}

\begin{document}
\maketitle
\begin{abstract}

Variational autoencoders (VAEs) are fundamental for generative modeling and image reconstruction, yet their performance often struggles to maintain high fidelity in reconstructions. This study introduces a hybrid model, quantum variational autoencoder (Q-VAE), which integrates quantum encoding within the encoder while utilizing fully connected layers to extract meaningful representations. The decoder uses transposed convolution layers for up-sampling. The Q-VAE is evaluated against the classical VAE and the classical direct-passing VAE, which utilizes windowed pooling filters. Results on the MNIST and USPS datasets demonstrate that Q-VAE consistently outperforms classical approaches, achieving lower Fréchet inception distance scores, thereby indicating superior image fidelity and enhanced reconstruction quality. These findings highlight the potential of Q-VAE for high-quality synthetic data generation and improved image reconstruction in generative models.

\end{abstract}

\section{Introduction}
Variational encoders (VAEs) are widely used in generative modeling and image reconstruction, providing an efficient framework to learn meaningful latent representations \cite{Kingma2013}. VAEs employ an encoder-decoder structure, where the encoder maps input data into a probabilistic latent space, and the decoder reconstructs the input from sampled latent variables. Fidelity in image reconstruction and generation refers to how well the generated images match the original images in terms of structural accuracy \cite{Duminil2024}, detail preservation \cite{xie2016detail}, and perceptual similarity \cite{mcnamara2006exploring}. However, classical VAEs (C-VAEs) often struggle to generate high-fidelity images due to inherent limitations in their latent representation \cite{yang2024}. Additionally, the decoder's reliance on pixel-wise reconstruction loss leads to blurry and over-smoothed images, restricting their effectiveness in capturing complex data distributions \cite{Khan2018-}.

Recent advancements in quantum computing offer promising avenues for overcoming these challenges by exploiting quantum phenomena such as superposition and entanglement. These advances in quantum computing have introduced promising approaches to enhance deep generative models by leveraging quantum parallelism and superposition \cite{Benedetti2019}. The integration of quantum circuits into neural networks, particularly within the VAE encoder, has shown potential to improve latent space representations, leading to enhanced feature extraction and more accurate reconstructions \cite{Ahmed2021}. Quantum neural networks represent an example of such hybrid models, which leverage quantum operations to better capture the underlying data structure, making them especially well suited for high-dimensional image tasks \cite{Senokosov2024}. 

This study presents a hybrid model referred to quantum VAE (Q-VAE). Q-VAE combines the quantum down sampling filter, which performs image resolution reduction and quantum encoding, and VAE. This is a novel approach that 
integrates quantum transformations within the encoder while utilizing fully connected layers for feature extraction. The decoder employs transposed convolution layers for upsampling. Q-VAE is evaluated against C-VAE and classical direct passing VAE (CDP-VAE), which incorporates windowed pooling filters (resolution reduction). Since quantum computing, especially its integration into deep learning models, is still in the early stages, focusing on simpler datasets allows for a more controlled environment where we can better understand the impact of quantum techniques on model performance. Furthermore, the datasets used in this study, such as MNIST \cite{deng2012mnist} and USPS \cite{usps_dataset} (with image resolutions of 28x28 and 16x16 pixels, respectively), are relatively small in size, which is advantageous when exploring the potential of quantum circuits for feature extraction and latent space representation. Their reduced resolution helps reduce the computational burden compared to higher-resolution images. To ensure a fair evaluation, the MNIST dataset is first down-sampled from 28x28 to 16x16 pixels, and then, like the USPS dataset, it is up-sampled to 32x32 pixels before being used as input to assess our model's performance. By focusing on grayscale images, we reduce the complexity of the problem, allowing for clearer insights into how quantum techniques can improve feature extraction and image reconstruction, particularly for simpler tasks like recognizing digits. Once the approach is successfully validated on these datasets, the methods can be extended to more complex and higher-resolution data, such as color images or higher resolutions.

Quantitative analysis has been performed to compare the effectiveness of the model using common performance metrics, such as the Fréchet Inception Distance (FID) and the Mean Squared Error (MSE), which are essential to assess the quality of generated images and the representation of the latent space \cite{yu2021fid, wang2002}. Thus, by leveraging these datasets, we can benchmark the Q-VAE against traditional models, to explore that if quantum component brings measurable improvements to the generative process.
In summary the following contributions have been made in this paper:
\begin{itemize}
    \item Introduces a new model, Q-VAE, where quantum encoding is applied only in the encoder, simplifying the model and improving the feature extraction capabilities.
    \item Compares Q-VAE with the CDP-VAE, highlighting performance differences due to the inclusion of quantum encoding in the encoder.
    \item  Avoids quantum entanglements or layers in the encoder, focusing solely on an efficient quantum encoding circuit for latent space encoding.
    \item  Minimizes quantum-trainable parameters, leading to more efficient model training and better utilization of quantum resources.
    \item  Evaluate whether quantum encoding improves the generative quality and  the latent space representation compared to classical methods.
    \item  This study explores the effects of upscaling images to 32×32 on model performance, examining how resolution influences feature extraction and representation.
\end{itemize}

The structure of this paper is as follows. Section 2 provides a review of related work on classical VAEs and their integration with quantum computing. Section 3 presents the background on VAEs. In Section 4, the methodology for the proposed Q-VAE model is introduced. Section 5 outlines the numerical simulations conducted, while Section 6 discusses the performance analysis. Finally, Section 7 concludes the paper, summarizing the key findings and suggesting potential avenues for future research.

\section{Related Research}

In recent years, VAEs have garnered significant attention in generative modeling, especially for image reconstruction tasks \cite{Kingma2013}. 
Traditional VAEs are valued for encoding data into a structured latent space and reconstructing input approximations, driving research efforts to enhance their architectures for improved image quality and feature extraction 
\cite{Mak2023, Neloy2024, Elbattah2021}. Several studies have focused on enhancing VAE performance using convolutional architectures \cite{Wang2024, Dai2019, Hou2017, Yang2017}. Gatopoulos et al. introduced convolutional VAEs to better capture image features, surpassing the performance of fully connected architectures in medium-resolution data sets such as CIFAR-10 and ImageNet32 \cite{Gatopoulos2020}. 
Furthermore, combining VAE with adversarial training, such as in Adversarial Autoencoders (AAE) \cite{Makhzani2015}, incorporates discriminator networks to enhance the visual quality of generated images, and improve machine learning applications. 
\subsection{Quantum Generative Modeling}
Quantum Machine Learning (QML) has emerged as a promising field, with studies exploring the integration of quantum circuits into generative models. Quantum GANs (QGANs) leverage quantum computing to enhance the data representation and image generation capabilities of GANs \cite{Ngo2023, Huang2021, Dallaire-Demers2018, niu2022entangling}. Quantum VAE \cite{Rocchetto2018, Gao2020, Gircha2021, Wang2024r, Luchnikov2019, Bhupati2024} on the other hand, is based on the foundational work by Khoshaman et al., 
who proposed a quantum VAE where the latent generative process is modeled as a quantum Boltzmann machine (QBM). Their approach demonstrates that quantum VAE can be trained end-to-end using a quantum lower bound to the variational log-likelihood, achieving state-of-the-art performance on the MNIST dataset \cite{Khoshaman2018}. They have used log-likelihood and Evidence Lower Bound (ELBO) as the measurement and have shown quantum VAE confidence level for the numerical results is smaller than ±0.2 in all cases. Gircha et al. further advanced the quantum VAE architecture by incorporating quantum circuits in both the encoder, which utilizes multiple rotations, and the decoder \cite{Gircha2021}. These quantum VAE models primarily focus on leveraging quantum methods in the latent space representation, enhancing the generative capabilities of VAEs. 

This paper addresses a gap in the literature where quantum techniques are primarily used in hybrid models for the encoder, while the decoder remains classical. Our proposed model is unique in that it leverages quantum encoding with a single rotation gate and measurement solely in the encoder, distinguishing it from other approaches.

\section{Background}
\subsection{Variational Auto-encoder}

VAEs are powerful generative models in machine learning, built on a probabilistic framework. The encoder network maps input data \( x \) to a latent distribution \( q_{\phi}(z|x) \), while the decoder network reconstructs the original data from the latent variables \( z \). Similar to other variational methods, VAEs are optimized using the ELBO as their objective function. Kingma et al. (2019) introduced a parametric inference model \( q_{\phi}(z|x) \), also referred to as an encoder or recognition model. The parameters of this inference model, denoted as \( \phi \), are known as the variational parameters. The decoder, with parameters \( \theta \), is responsible for generating the data from the latent variables \( z \). Thus, the parameters \( \phi \) and \( \theta \) together define the model, where \( \phi \) governs the inference process and \( \theta \) governs the generative process.\cite{Kingma2019}. 


Due to the non-negativity of the  Kullback-Leibler (KL) 
divergence between \( q_{\phi}(z|x) \) and \( p_{\theta}(z|x) \), ELBO serves as a lower bound on the logarithmic likelihood of the data. This relationship is expressed as:
\begin{equation}
L_{\theta, \phi}(x) = \log p_{\theta}(x) - D_{KL}(q_{\phi}(z | x) || p_{\theta}(z | x))
\end{equation}

Interestingly, the KL divergence \( D_{KL}(q_{\phi}(z | x) || p_{\theta}(z | x)) \) determines two key aspects: 
\begin{enumerate}
    \item By definition, it quantifies the divergence of the approximate posterior from the true posterior.
    \item It represents the gap between the ELBO \( L_{\theta, \phi}(x) \) and the marginal likelihood \( \log p_{\theta}(x) \), which is also known as the \textit{tightness of the bound}.
\end{enumerate}
The closer \( q_{\phi}(z | x) \) approximates the true posterior \( p_{\theta}(z | x) \) in terms of KL divergence, the smaller this gap becomes, leading to a tighter bound. The reconstruction term encourages the model to generate data points that are similar to the observed data, while the KL divergence regularizes the learned latent variables by pushing the approximate posterior towards the prior distribution.
Although VAEs have been successful in many domains, including image generation and anomaly detection, they often face significant challenges when applied to high-resolution data. This issue can arise because the latent space is often not expressive enough to model the intricate relationships. \cite{Hu2024}.

\subsection{Encoder}
A classical VAE encoder is typically implemented as a fully connected network or convolutional neural network (CNN) 
with the aim of mapping the input data to a lower-resolution latent space. The primary objective of the encoder is to process the input data \( x \), transform it through a series of hidden layers, and output two parameters: the mean \( \mu \) and the log variance \( \log(\sigma^2) \), which are used for variational inference \cite{Pinheiro2021}.
In the case of image data, the input image is first reshaped into a vector and passed through multiple fully connected layers. The transformations performed by these layers help extract important features from the input. For example, given an input image of dimension 1024 (flattened from a \( 32 \times 32 \) image), the encoder progressively reduces the dimension, often mapping the input image \( x_0 \) into a lower-dimension space \( x_1 \). 

\subsubsection{Fully Connected Layers in Encoders}

Fully connected layers in an encoder are crucial for transforming input data into meaningful representations. They work by connecting every neuron in one layer to every neuron in the next, allowing the model to learn complex relationships between features. In the encoder, these layers typically perform dimensionality reduction, extracting the most relevant information from the input and mapping it to a compact latent space. The use of activation functions like ReLU or Sigmoid introduces non-linearity, enabling the encoder to capture intricate patterns. This learning of complex, non-linear relationships and improving performance in capturing patterns make fully connected layers essential for building efficient and powerful encoders. This approach allows the model to better capture spatial dependencies while still benefiting from the representational power of fully connected layers.

\subsubsection{Windowing and Pooling for Resolution Reduction}

Fixed pooling techniques, such as max pooling and average pooling, are particularly effective in encoding processes where spatial reduction is necessary. By consistently applying the same pooling operation across the input, the model can reduce dimensionality, improve efficiency, and retain key features that will be crucial for later processing.
In max pooling, the maximum value in a small region (or window) of the input image is selected, reducing the image's resolution while preserving the most significant features in each region. Average pooling works similarly, but instead computes the average value of each window. These pooling operations help to reduce computational complexity while retaining essential information for later stages of the model.

\subsection{Latent Space Encoding} 
\label{sec:headings}

Finally, the encoder outputs the mean \( \mu \) and the log variance \( \log(\sigma^2) \), which parameterize the variational distribution. These values are computed using fully connected layers (fc) as follows:

\begin{equation}
\mu = \text{fc}_{\mu}(h_2),
\end{equation}
\begin{equation}
\log(\sigma^2) = \text{fc}_{\log\_var}(h_2),
\end{equation}
where \( \mu \) and \( \log(\sigma^2) \) are vectors of dimension of the latent space, while \( h_2 \) is a vector or matrix resulting from a series of transformations applied to the input data, and it is used as the input to the fully connected layers that output the parameters of the variational distribution.
.



\subsection{Decoder} 
\label{sec:decoder}

The decoder in a VAE is designed to transform the lower-resolution latent variable \( z \), produced by the encoder, back into a high-dimensional data space, reconstructing the input data. The decoder typically consists of multiple layers, including fully connected and transposed convolution layers, which progressively reconstruct the input data's spatial and feature dimensions.
The core function of the decoder is to ensure that the reconstructed data match closely the original input, enabling the model to learn both the global structure and the local details of the data. As with the encoder, the decoder is trained to minimize the reconstruction error, often using a loss function that combines both the likelihood of the data given the latent variables and a regularization term that encourages smoothness in the latent-space representation.

\subsubsection{Latent Variable Transformation}

The decoding process begins by passing the latent variable \( z \) through a fully connected layer to map it to a higher-dimensional feature space. This transformation creates a tensor that matches the expected dimensions for the next layers in the decoder, such as the transposed convolution layers. The fully connected layer enables the decoder to effectively learn a mapping from the compact latent space back to the data space, ensuring that relevant information is maintained during this transformation.

\subsubsection{Transposed Convolutions}

Once the latent variable is transformed into the higher-dimensional space, it is passed through a series of transposed convolution layers. These layers up-sample the tensor, progressively restoring the original spatial dimensions of the input data. Each transposed convolution layer learns to refine the spatial resolution of the image while preserving important features.
Transposed convolution layers have been shown to be highly effective in generating high-resolution output from low-resolution latent representations. They are capable of learning hierarchical features of the data, capturing both local details and global structure during the reconstruction process. By applying non-linear activation functions, such as ReLU, between the transposed convolution layers, the model introduces the necessary complexity to capture intricate patterns in the data.

\subsubsection{Reconstruction}

The final stage of the decoding process involves the production of the reconstructed image. This is typically done by passing the output of the last transposed convolution layer through a sigmoid activation function to ensure that the pixel values lie within the valid range (e.g., between 0 and 1 for normalized image data). The output of this step is the reconstructed image, which should ideally match the original input data.

\section{Proposed Methodology}

In this work, we introduce a novel down-sampling filter for Q-VAE 
that relies exclusively on quantum encoding for feature extraction. The paper investigates three distinct architectures for VAEs, each employing different encoding mechanisms to evaluate their feature extraction capabilities: C-VAE, CDP-VAE and Q-VAE. The framework of these models is presented in Figure 1. The goal of this comparison is to assess the effectiveness of quantum computing, classical neural networks, and simplified encoding strategies in the capture and processing of data for subsequent reconstruction. Although each model shares a common decoder architecture, the encoders differ in their approach, enabling the study of how each influences the encoding process and the quality of the reconstructed output. 

\begin{figure}[h!]
    \centering
    \includegraphics[width=400pt,height=50pc]{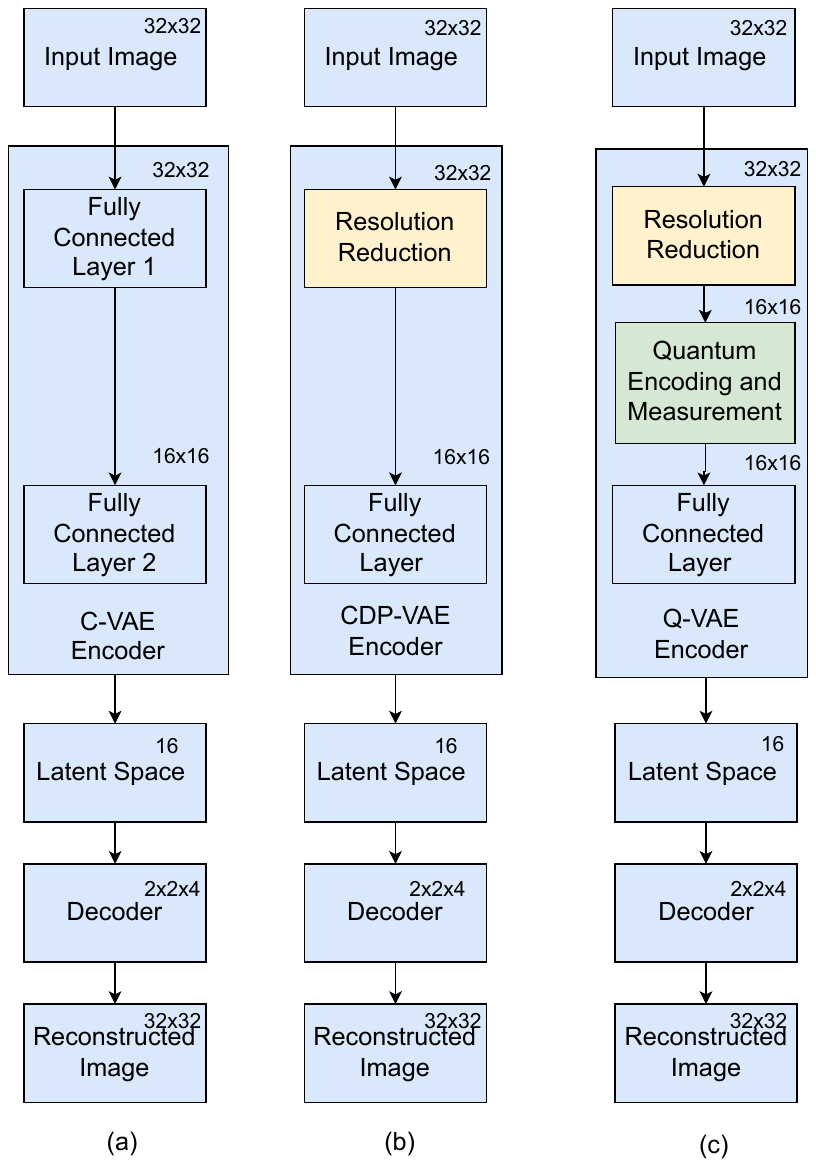}
    \caption{Flowchart for (a) C-VAE, (b) CDP-VAE and (c) Q-VAE.} 
    \label{fig1}
\end{figure}

\subsection{Classical VAE (C-VAE) Encoder}

C-VAE uses a traditional fully connected neural network as the encoder, which processes the input data through a series of hidden layers to output the mean (\( \mu \)) and log variance (\( \log(\sigma^2) \)) of the latent space distribution. This distribution is then sampled to obtain the latent variable \( z \), which represents the encoded data in a lower-resolution space. While the C-VAE relies entirely on classical neural networks for encoding, it remains highly effective in modeling data distributions and is widely used in practical applications. The decoder,which employs layers for up-sampling, reconstructs the input from this low-dimension latent space, generating an output image or data structure that approximates the original input, also shown in Figure~\ref{fig1} (a).

\subsection{Classical Direct Passing (CDP-VAE) Encoder} 

CDP-VAE adopts a more simplified approach to encoding by using windowing pooling for resolution reduction as shown in Figure~\ref{fig1} (b). In this method, the input data (e.g., an image) is divided into \( 2 \times 2 \) pixel windows, and only the first pixel of each window is retained, discarding the other three. This process reduces the resolution of the input data while preserving significant features. The reduced input is then passed through a fully connected layer for further encoding into the latent space. The CDP-VAE encoder is particularly useful for computationally constrained environments, where reducing the complexity of the encoding process can lead to faster computations while still capturing essential information for reconstruction. For instance, if the input image \( x \) is of dimension \( 32 \times 32 \), this resolution reduction process reduces it to \( 16 \times 16 \).



\subsection{Quantum VAE (Q-VAE) Encoder}

Q-VAE integrates quantum computing into the encoding process, adding in the CDP-VAE  as shown in Figure~\ref{fig1} (c). The quantum encoder processes input data into a quantum state representation, utilizing quantum superposition to represent intricate relationships within the data.
Quantum encoding improves performance by transforming input data into quantum states, allowing the Q-VAE to apply rotations that facilitate efficient processing and capture richer, more intricate representations.
The quantum-enhanced feature extraction is combined with a classical decoder that uses transposed convolution to reconstruct the input data, combining the strengths of quantum processing for feature extraction with classical techniques for data reconstruction. 

\begin{figure}[h!]
    \centering
    \includegraphics[width=0.9\linewidth, height=4cm]{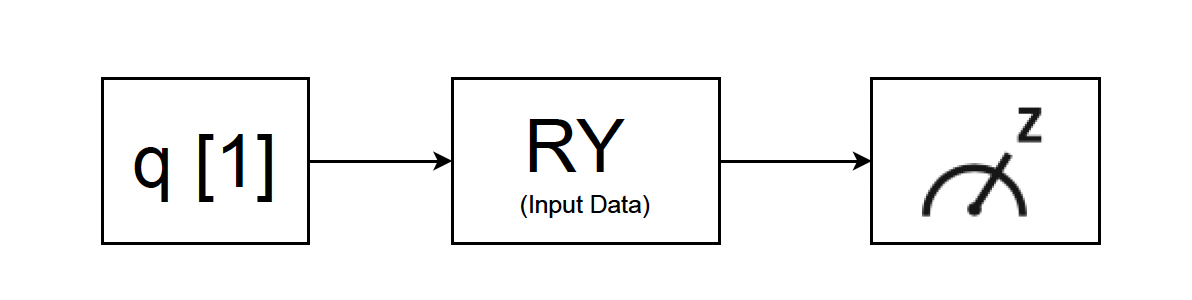}
    \caption{Proposed Quantum circuit for Quantum Variational Auto-encoder} 
    \label{fig3}
\end{figure}

The quantum encoding network, illustrated in Figure~\ref{fig3}, uses a quantum encoding designed to process 16x16 pixel images that is the output of windowing pooling layer. 

\begin{equation}
R_Y(x_1) = \exp\left(-i \frac{x_1}{2} Y\right),
\end{equation}
where \( R_Y \) is the rotation gate around the \( Y \)-axis.
The rotation gate \( R_Y(\theta) \) can be expressed in terms of sine and cosine as:

\begin{equation}
R_Y(\theta) = \exp\left(-i \frac{\theta}{2} Y\right) = \cos\left(\frac{\theta}{2}\right) I - i \sin\left(\frac{\theta}{2}\right) Y
\end{equation}

where \( I \) is the identity matrix and \( Y \) is the Pauli \( Y \)-matrix. The explicit matrix form is:

\begin{equation}
R_Y(\theta) = \begin{pmatrix} \cos\left(\frac{\theta}{2}\right) & -\sin\left(\frac{\theta}{2}\right) \\ \sin\left(\frac{\theta}{2}\right) & \cos\left(\frac{\theta}{2}\right) \end{pmatrix}
\end{equation}

For the rotation gate applied to the first pixel, \( R_Y(x_1) \) becomes:

\begin{equation}
R_Y(x_1) = \begin{pmatrix} \cos\left(\frac{x_1}{2}\right) & -\sin\left(\frac{x_1}{2}\right) \\ \sin\left(\frac{x_1}{2}\right) & \cos\left(\frac{x_1}{2}\right) \end{pmatrix}
\end{equation}
The Pauli-\( Z \) operator is applied to measure the quantum state. The Pauli-\( Z \) measurement flips the qubit’s state depending on the computational basis, effectively capturing the binary features of the image.

Mathematically, the Pauli-\( Z \) operator acts on the quantum state \( \left| \psi \right\rangle \) as follows:
where \( Z \) flips the phase of the qubit in the \( \left| 1 \right\rangle \) state, while leaving the \( \left| 0 \right\rangle \) state unchanged. 
This operation effectively translates quantum information into meaningful latent variables for the model. The resulting latent variables \( z \) are given by:
\begin{equation}
z = \text{Measurement}(\left| \psi \right\rangle, Z)
\end{equation}
The measurement on qubit 
yields the compressed representation of the input. This procedure reduces the dimension of the input image from 1024 to 256 measured values. After the quantum circuit processes the input image and extracts quantum features, the output of the quantum circuit, as shown in Figure~\ref{fig1} (c), is passed to fully connected layer. This transition enables the integration of quantum-encoded features with classical deep learning architectures.




\subsection{Shared Decoder Architecture} 

Despite the differences in their encoding mechanisms, all three models, Q-VAE, C-VAE, and CDP-VAE, share a common decoder architecture. This decoder is responsible for reconstructing the input data from the latent space representation. The decoder begins by transforming the latent variable (z) into a medium-resolution space through fully connected layers. 
Then a series of transposed convolution (deconvolution) layers are used to progressively up-sample the data, restoring the original spatial dimensions. Finally, the output is passed through a sigmoid activation function to ensure that the pixel values are within the valid range (e.g,, between 0 and 1 for image data). This shared decoder allows for a fair comparison of the encoding techniques, with the focus on how the different encoders influence the quality of the reconstructed image.

\section{Numerical Simulation}\label{sec4}

To assess the performance of our proposed model, we first need to preprocess the input image data. The preprocessing steps are described in the following sections.

\subsection{USPS Dataset Pre-Processing}

USPS dataset input image size resolution is 16x16. The USPS images are resized to \(32 \times 32\) using the same resizing function as was done for MNIST.

\begin{figure}[h!]
    \centering
    \begin{subfigure}[b]{0.45\textwidth}
        \centering
        \includegraphics[width=\textwidth,height=150pt]{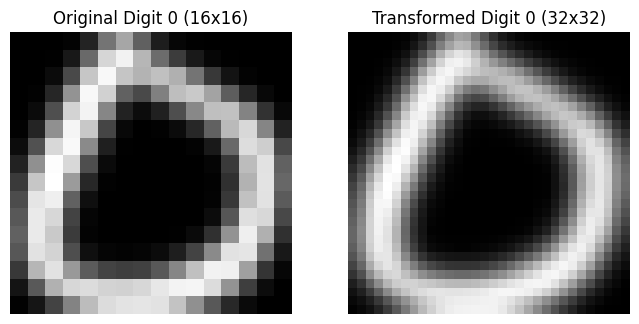}
        \label{fig:image1}
    \end{subfigure}
    \hfill
    \begin{subfigure}[b]{0.45\textwidth}
        \centering
        \includegraphics[width=\textwidth,height=150pt]{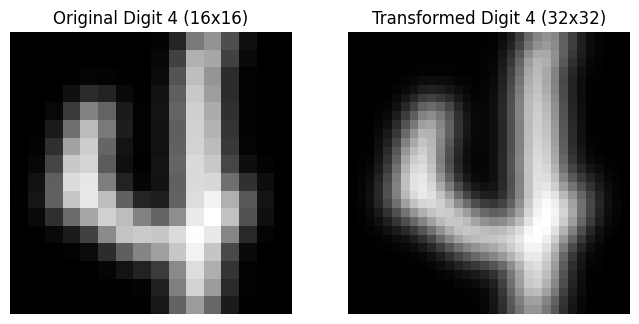}
        \label{fig:image2}
    \end{subfigure}

    \caption{USPS Image transformed from 16x16 to 32x32.}
    \label{fig7:multiple_images}
\end{figure}
The interpolation method used during resizing adjusts the pixel values to fit the larger image grid, enabling the model to be tested with standard image input sizes, such as 32x32. USPS image conversion is shown in Figure~\ref{fig7:multiple_images}.

\subsection{MNIST Dataset Pre-Processing}

As observed in USPS, We utilized MNIST to determine whether the performance improvements observed in USPS could be replicated with MNIST. To enhance the model’s ability to generalize across different image sizes, we perform the following preprocessing steps:

\begin{enumerate}
    \item \textbf{Cropping:} Initially, the center of each image is cropped to remove any surrounding whitespace, reducing the image size from \(28 \times 28\) pixels to a smaller region that more tightly focuses on the handwritten digit (important features). The cropping ensures that the model receives more relevant features of the digits.
    
    \item \textbf{Resizing to \(16 \times 16\):} After cropping, the image is resized to \(16 \times 16\) pixels. This step reduces the image resolution, making it easier for the model to process while still retaining key features of the digit. The resizing step uses an interpolation technique that preserves the overall structure of the digit. This process will help us to get same resolution dataset like USPS to see if MNIST also show improvement or not.
    
    \item \textbf{Resizing to \(32 \times 32\):} Finally, the image is resized to \(32 \times 32\) pixels. This larger size is used for feeding the images into the neural network model, which is optimized to process medium resolution images.This study investigates how resizing adjusts the image resolution and its impact on the model’s ability to process finer details for reconstruction and classification. 
    
\end{enumerate}

\begin{figure}[h!]
    \centering
    \begin{subfigure}[b]{0.45\textwidth}
        \centering
        \includegraphics[width=\textwidth,height=150pt]{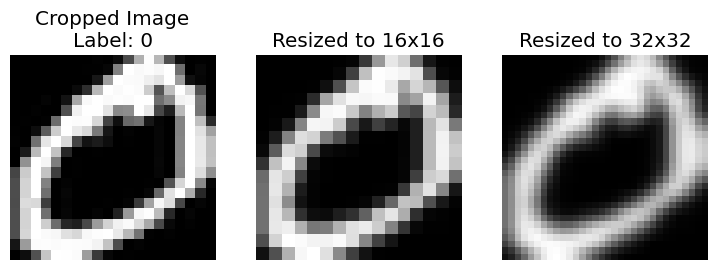}
        \label{fig:image1}
    \end{subfigure}
    \hfill
    \begin{subfigure}[b]{0.45\textwidth}
        \centering
        \includegraphics[width=\textwidth,height=150pt]{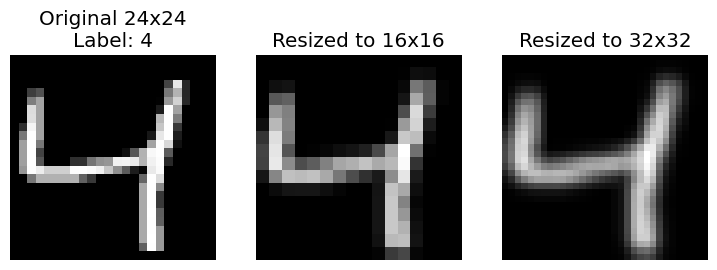}
        \label{fig:image2}
    \end{subfigure}

    \caption{MNIST Image cropped for digit then transformed to 16x16 to 32x32.}
    \label{fig6:multiple_images}
\end{figure}

Figure~\ref{fig6:multiple_images} shows the transformation of MNIST images as processed according to the above procedure.  This resizing step ensures compatibility with the input size of the model while retaining the features necessary for effective representation learning.

\subsection{Experimental Setup}

In this section, we describe the hyper-parameters and settings used for training the models. We are implementing our algorithm on Pennylane quantum simulator. The code is available at https://github.com/RRFar/Quantum-Downsampling-Filter. Hyper-parameters to run the algorithm were carefully selected to ensure stable training while optimizing performance for all models.

\subsubsection{Hyper-parameters}

In our training process, several key hyper-parameters were used to ensure efficient learning and model convergence. The learning rate was set to \(0.001\) for all  the models. This learning rate controls how much the model's weights are updated in response to the computed gradients, providing a stable training process while allowing sufficient progress over the course of the training. The models were trained for 200 epochs, with each epoch representing one full pass of the data set through the model. This duration was chosen to ensure that the models had enough time to learn the underlying features of the data, while regular monitoring of the validation loss helped prevent over-fitting. A batch size of \(400\) was used, which means that \(400\) images were processed together before updating the model parameters. This batch size was selected to balance computational efficiency and the stability of gradient estimation, helping smooth gradient updates and ensuring more stable training.
For optimization, we used the Adam optimizer, which is known for its adaptive learning rate that adjusts based on the first and second moments of the gradients. The Adam optimizer is particularly effective for models with many parameters and large datasets. The loss function used was the binary cross-entropy loss (BCELoss), which ensures that the individual losses across all pixels are summed rather than averaged. This choice emphasizes the total error in the reconstruction of the images. These images allowed us to visually assess how well the models were reconstructing the images as training progressed. These configurations were chosen to ensure effective model training, allowing the models to learn from the data efficiently and perform well in both reconstruction and evaluation tasks.

\section{FID and MSE Score Performance Evaluation and Discussion }\label{sec4.1}

\begin{figure}[h!]
    \centering
   \includegraphics[width=350pt,height=22pc]{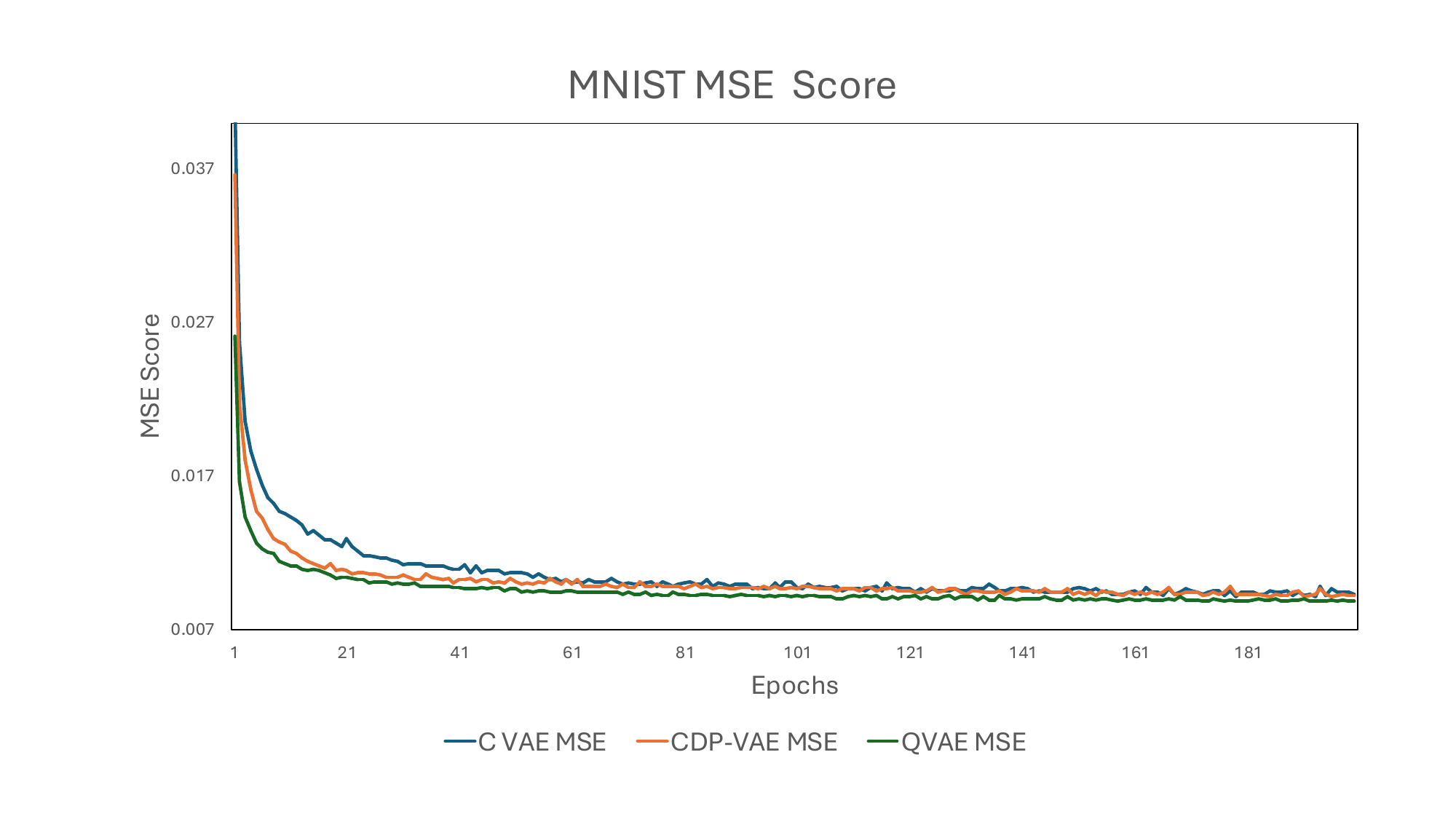}
    \caption{MNIST MSE score for all models}
    \label{fig7}
\end{figure}

\begin{figure}[h!]
    \centering
    \includegraphics[width=350pt,height=22pc]{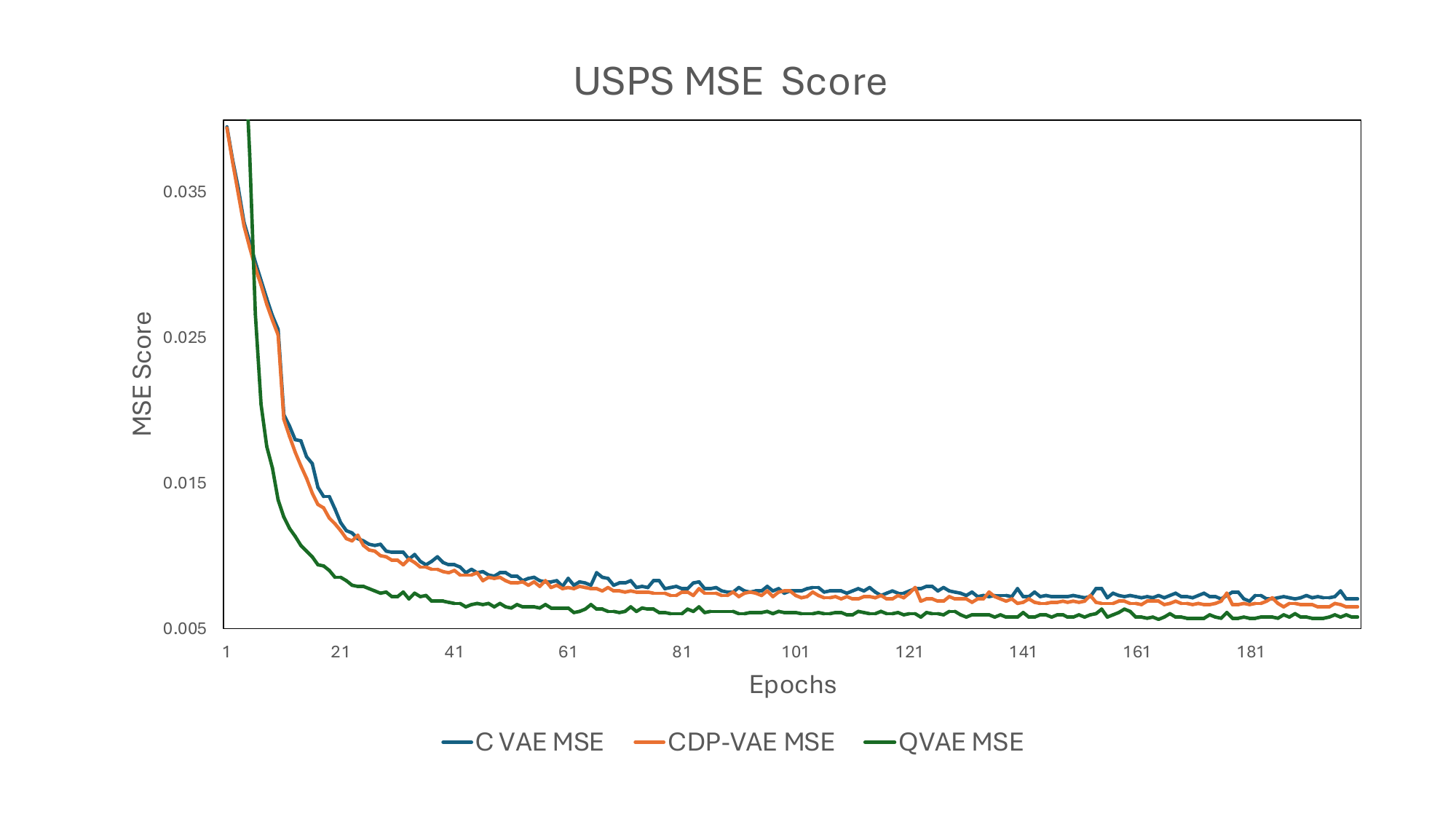}
    \caption{USPS MSE score for all models}
    \label{fig8}
\end{figure}

To evaluate the performance of the proposed models, we used a combination of several key metrics: FID score \cite{yu2021fid}, training and testing loss, and MSE \cite{wang2002}. 
Each of these metrics provides distinct insights into the model’s ability to generate better-quality images and generalize to new data:
FID measures the similarity between the distribution of real and generated images. A lower FID score indicates that the generated images are closer in distribution to the real images, reflecting higher image fidelity and quality. The FID score is particularly useful for assessing the realism of generated images in generative models. 
MSE measures the average squared difference between the generated and real images. A lower MSE indicates that the generated images are closer to the real images in terms of pixel-wise accuracy. This metric is commonly used to evaluate the reconstruction quality in generative models, with lower values indicating better performance. MNIST and USPS MSE results are shown in Figure~\ref{fig7} and Figure~\ref{fig8} respectively. 


\begin{figure}[h!]
    \centering
   \includegraphics[width=350pt,height=22pc]{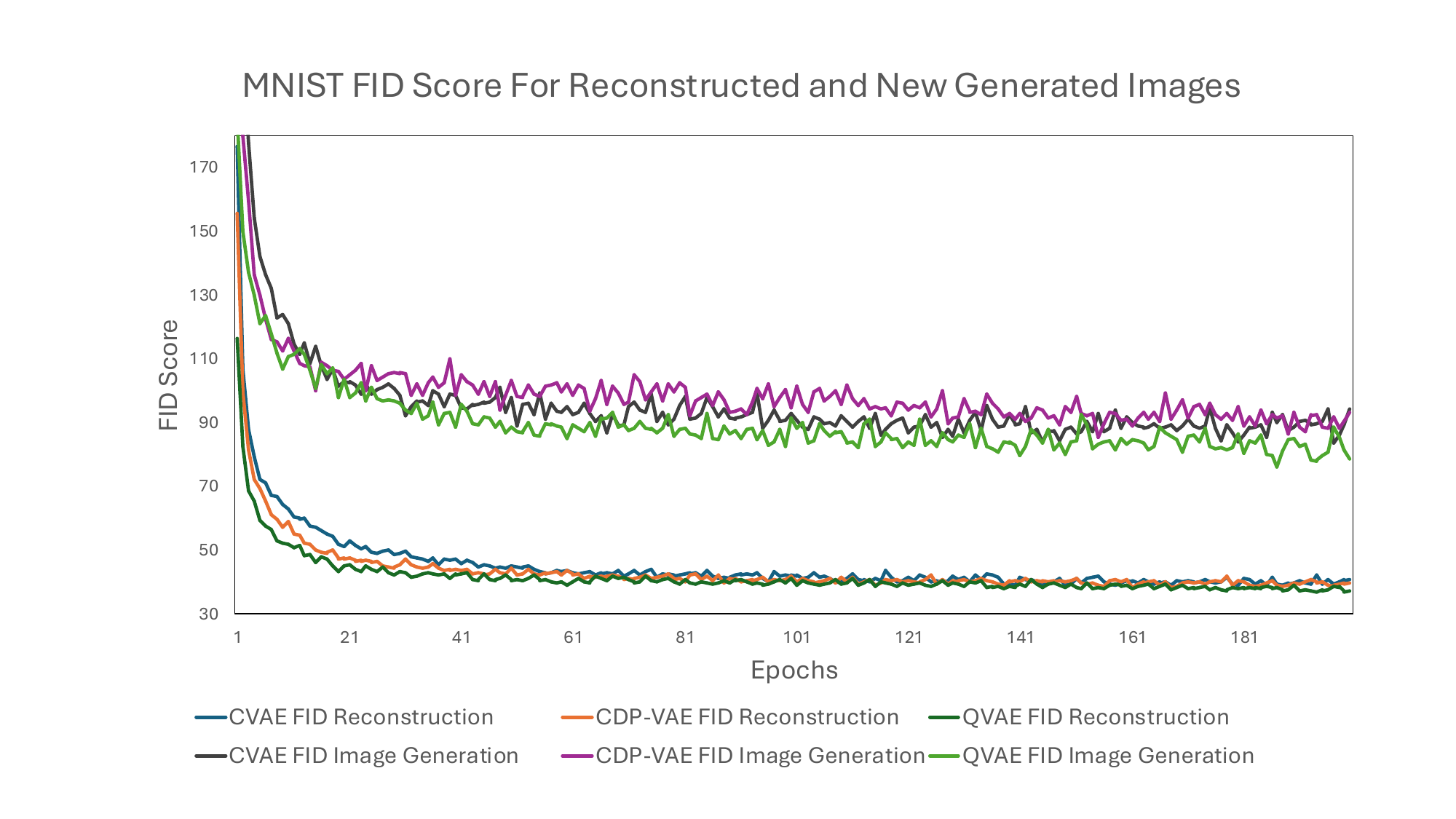}
    \caption{MNIST FID score for image reconstruction and image generation.}
    \label{fig9}
\end{figure}

By combining these metrics, we gain a comprehensive understanding of how well our models are performing in terms of both image quality and generalization ability. The results are shown in Table 1 below:

\begin{table}[ht]
\caption{Experimental results}
\centering
\begin{tabular}{|l|c|c|c|}
    \hline
    Datasets & FID score for image reconstruction & FID score for image generation & MSE \\ \hline
    C-MNIST  & 40.7 & 94.4 & 0.0093 \\ \hline
    CDP-MNIST & 39.7 & 93.3 & 0.0092 \\ \hline
    \textbf{Q-MNIST}  & \textbf{37.3} & \textbf{78.7} & \textbf{0.0088} \\ \hline
    C-USPS   & 50.4 & 73.1 & 0.0070 \\ \hline
    CDP-USPS & 42.9 & 66.1 & 0.0065 \\ \hline
    \textbf{Q-USPS}   & \textbf{38.5} & \textbf{57.6} & \textbf{0.0058} \\ \hline\end{tabular}
\label{tab2}
\end{table}
The FID score was computed for 1) images reconstructed from the input images and 2) images generated from random noise, giving us insight into how well the models perform in generating realistic images. 
The variation of FID score as a function of the number of epochs for MNIST is shown in Figure~\ref{fig9}.

For MNIST, the C-VAE achieves an FID score of 40.7, indicating a reasonable ability to reconstruct images. The Q-VAE, on the other hand, leverages quantum gates such as RY rotations and Pauli-Z measurements to encode the data, achieving a superior FID score of 37.3. This improved score suggests that the quantum model produces more faithful image reconstructions, capturing complex relationships in the data more effectively than the classical model. Finally, the CDP-VAE, demonstrates a slight improvement over the C-VAE with an FID score of 39.7.  Similar trends have been noticed for USPS where C-VAE achieved 50.4, CDP-VAE scored 42.9 and Q-VAE FID score for reconstruction as 38.5. Although this approach is less complex, it still lacks the advanced feature extraction capabilities of the Q-VAE, which ultimately leads to lower reconstruction fidelity.

For the USPS dataset image generation, the Q-VAE achieved an FID score of 57.6, while MNIST achieved 78.7 
indicating better-quality image reconstructions and strong performance in terms of both fidelity and distribution similarity to the original dataset. While for MNIST C-VAE and CDP-VAE scored 94.4 and 93.3 respectively. Similar trends have been seen for USPS where C-VAE and CDP-VAE scored 73.1 and 66.1 respectively. This result highlights the effectiveness of the Q-VAE in handling simpler grayscale datasets, where it demonstrates feature extraction and reconstruction capabilities.

\begin{figure}[h!]
    \centering
   \includegraphics[width=350pt,height=20pc]{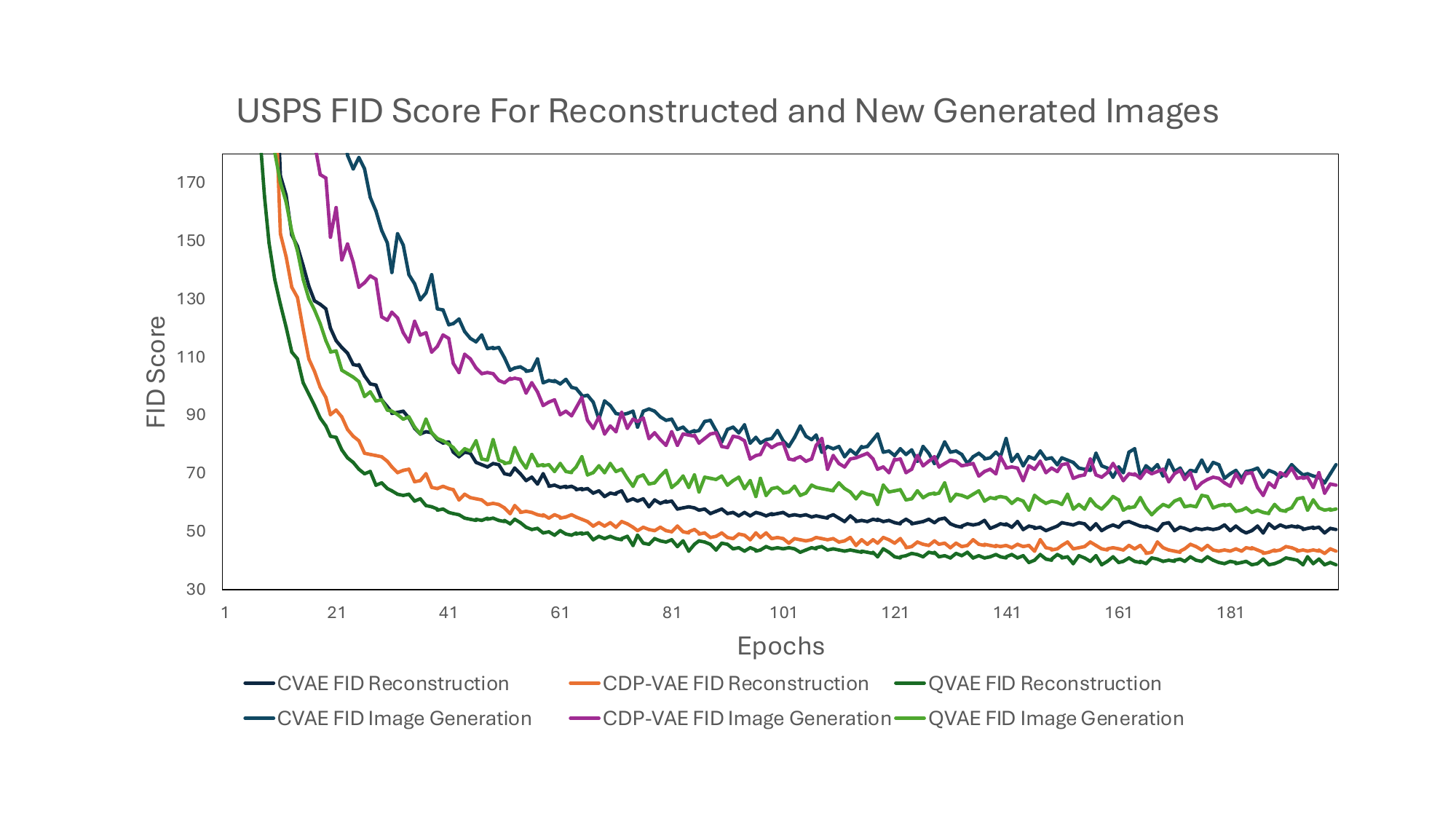}
    \caption{ USPS FID score for image reconstruction and image generation.}
    \label{fig10}
\end{figure}

The CDP-VAE is a VAE where, instead of using an encoder-decoder architecture with learned transformations, the input data is directly passed through with minimal transformation using windowing/ pooling. Despite its simplicity, the CDP-VAE still achieved relatively good performance on the USPS dataset, with an FID score of 66.1 as shown in Table \ref{tab2}. Figure~\ref{fig10} further illustrates that the Q-VAE consistently achieves a lower FID score for reconstruction compared to both CDP-VAE and C-VAE, indicating its superior ability to generate high-fidelity reconstructions.

\subsection{Latent Space Visualization} 

\begin{figure}[h!]
    \centering
   \includegraphics[width=250pt,height=16pc]{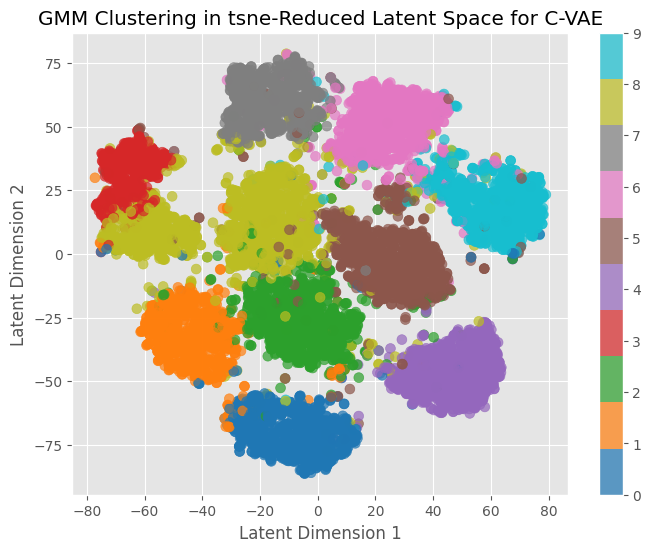}
    \caption{Latent space for C-VAE}
    \label{fig11}
\end{figure}

\begin{figure}[h!]
    \centering
   \includegraphics[width=250pt,height=16pc]{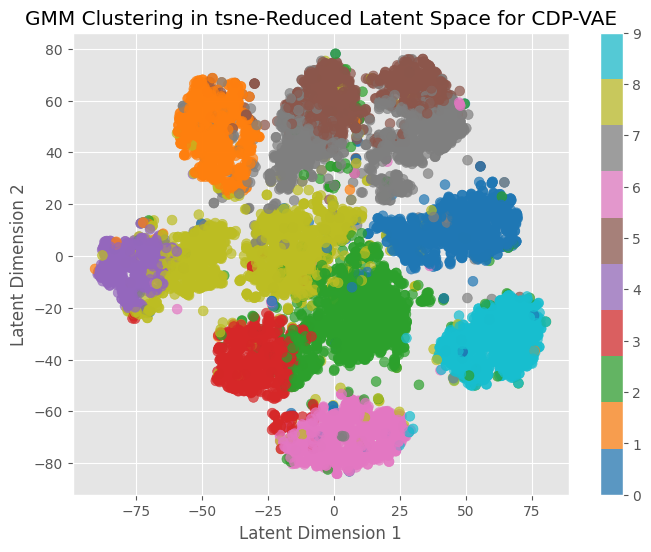}
    \caption{Latent space for CDP-VAE}
    \label{fig12}
\end{figure}

\begin{figure}[h!]
    \centering
   \includegraphics[width=250pt,height=16pc]{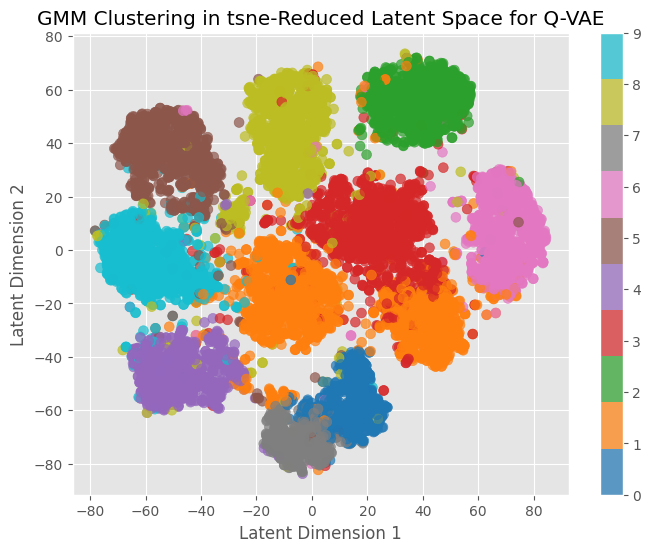}
    \caption{Latent space for Q-VAE}
    \label{fig13}
\end{figure}

To better understand the expressiveness of the model, the latent vector mapping is visualized using dimensionality reduction techniques such as t-SNE \cite{maaten2008tsne}, which shows the clustering structure in the latent space. Furthermore, Gaussian Mixture Model (GMM) clustering is used to assess how well the learnt representations can be grouped, providing information on the model's capacity to capture meaningful variations and structure in the data. The colours in the visualisation reflect different classes, with well-separated colours indicating efficient feature learning and overlapping colours indicating class similarities or ambiguity. Distinct clusters suggest that the model has learnt meaningful feature representations, but overlapping clusters could indicate insufficient separation in the latent space. By combining t-SNE and GMM clustering, we can assess whether the model effectively organises data points and captures key properties for classification or reconstruction. C-VAE, CDP-VAE and Q-VAE latent space has been visulatise in Figure ~\ref{fig11},~\ref{fig12} and ~\ref{fig13} respectively.


From the figures, it is evident that CDP-VAE and Q-VAE do not show a significant difference from C-VAE, but demonstrate competitive performance. For further investigation, we will analyze the reconstructed images and newly generated images in the next section.

\subsection{Reconstructed Images} 
\begin{figure}[h!]
    \centering
    \begin{subfigure}[b]{0.22\textwidth}
        \centering
        \includegraphics[width=\linewidth,height=250pt]{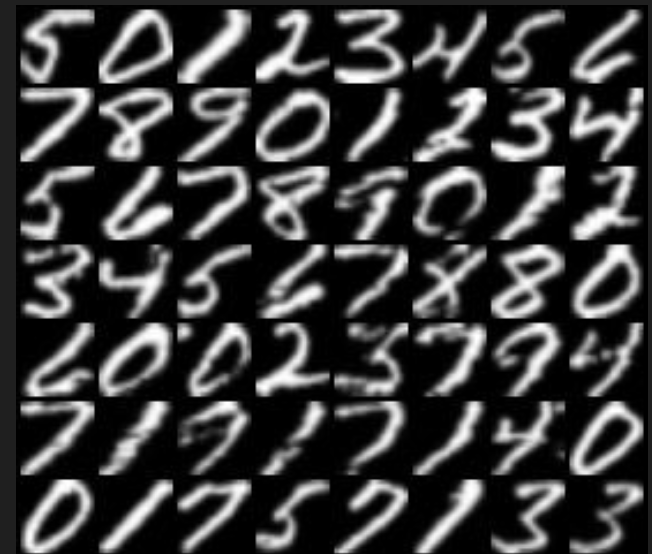}
    \caption{ Input Image }
        \label{fig:classical_reconstruction_1}
    \end{subfigure}    
    \hspace{0.02\textwidth} 
    \begin{subfigure}[b]{0.22\textwidth}
        \centering
        \includegraphics[width=\linewidth,height=250pt]{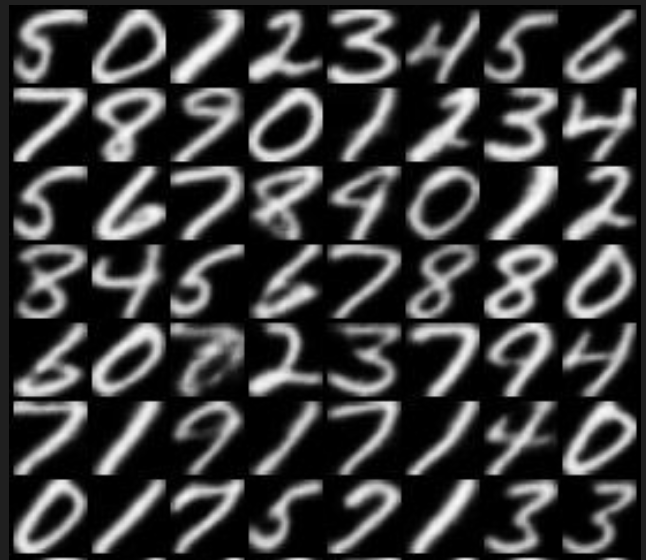}
        \caption{C-VAE}
        \label{fig:classical_reconstruction_1}
    \end{subfigure}    
    \hspace{0.02\textwidth} 
    \begin{subfigure}[b]{0.22\textwidth}
        \centering
        \includegraphics[width=\linewidth,height=250pt]{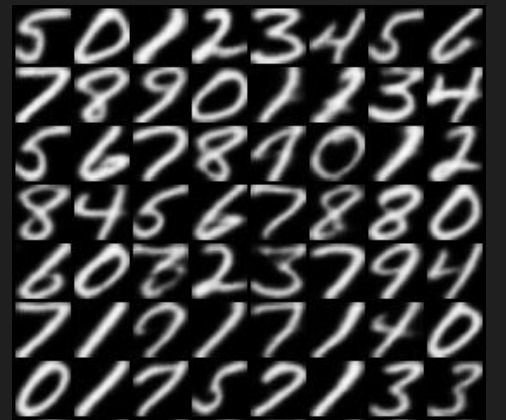} 
        \caption{CDP-VAE}
        \label{fig:classical_reconstruction_2}
    \end{subfigure}    
    \hspace{0.02\textwidth} 
    \begin{subfigure}[b]{0.22\textwidth}
        \centering
        \includegraphics[width=\linewidth,height=250pt]{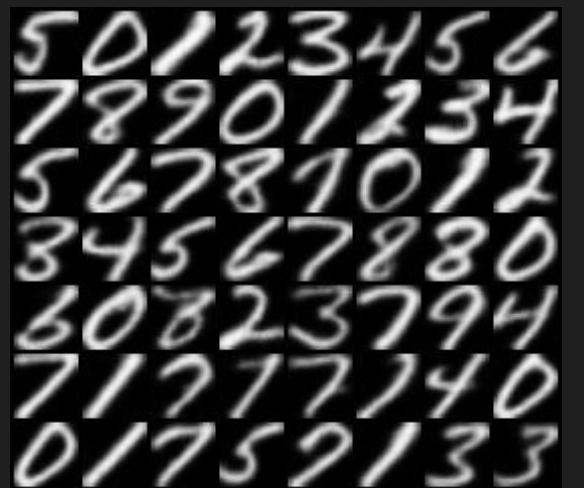} 
        \caption{Q-VAE}
        \label{fig9:quantum_reconstruction_2}
    \end{subfigure}

    \caption{Comparing classical and quantum reconstructed images. (a) MNIST input image (b) C-VAE image reconstructions, (c) CDP-VAE image reconstruction and (d) Q-VAE image reconstruction.}
    \label{fig:14four_images}
\end{figure}

Figure~\ref{fig:14four_images} displays reconstructed grayscale digit images from the MNIST dataset, arranged in a grid format. The input image is first downscaled from 28x28 to 16x16, and then upscaled to 32x32. This sequence reveals that some images lose details during resolution reduction, leading to inaccuracies in their reconstruction. For example, the digit '3' in the 4th row and 1st column is successfully reconstructed by Q-VAE, whereas the C-VAE and CDP-VAE struggle to generate a well-defined version. However, in the 5th row and 3rd column, the digit '0' shows noticeable noise due to the resolution reduction, making it difficult to reconstruct accurately. In this case, C-VAE incorrectly generates a '7', while Q-VAE produces a hybrid image that blends '7' and '0'. Most digits retain their recognizable shapes, but some exhibit blurriness or distortions as column 3, row 5 and '6' its difficult to recognise the image as '7' and '9', indicating challenges in capturing finer details during the reconstruction process. These issues may stem from factors such as limited latent space resolution's used for the model (i.e 16), preprocessing steps like resizing or normalization, or suboptimal model architecture. Interestingly, prior evaluations using metrics like the FID have demonstrated 
our proposed model Q-VAE, achieve better image quality on MNIST. 

On the MNIST dataset, the Q-VAE achieved the lowest MSE of 0.0088, demonstrating its ability to produce high-fidelity reconstructions compared to the other models. The quantum-based encoding method enables the Q-VAE to extract richer feature representations, leading to more accurate reconstructions. The quantum model's ability to capture complex dependencies between pixels allows it to outperform classical approaches, even when dealing with highly structured images like those in the MNIST dataset. The C-VAE, with an MSE of 0.0093, showed slightly higher error compared to the Q-VAE, indicating that while it performs well, its feature extraction capabilities are not as advanced as those of the quantum model. Despite its relatively low MSE, the classical model’s performance is constrained by the limitations of traditional neural network-based encoding and decoding methods, which cannot capture the full complexity of the data as efficiently as quantum circuits. The  CDP-VAE, with an MSE of 0.0092, performs little better then C-VAE but still didn't perform better then Q-VAE as shown in Figure~\ref{fig:14four_images}.

\begin{figure}[h!]
    \centering
    \begin{subfigure}[b]{0.22\textwidth}
        \centering
        \includegraphics[width=\linewidth,height=250pt]{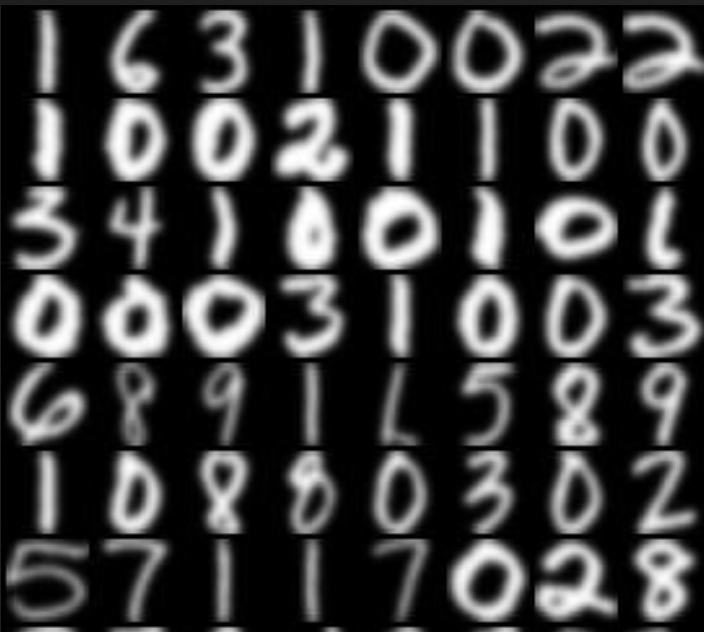}
    \caption{ Input Image }
        \label{fig:classical_reconstruction_1}
    \end{subfigure}    
    \hspace{0.02\textwidth} 
    \begin{subfigure}[b]{0.22\textwidth}
        \centering
        \includegraphics[width=\linewidth,height=250pt]{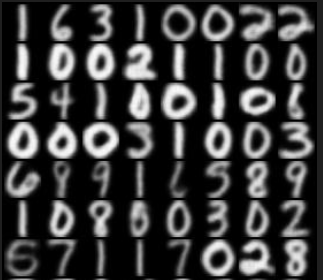}
        \caption{C-VAE}
        \label{fig:classical_reconstruction_1}
    \end{subfigure}    
    \hspace{0.02\textwidth} 
    \begin{subfigure}[b]{0.22\textwidth}
        \centering
        \includegraphics[width=\linewidth,height=250pt]{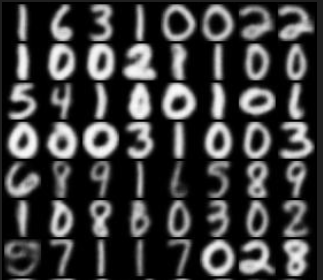} 
        \caption{CDP-VAE}
        \label{fig:classical_reconstruction_2}
    \end{subfigure}    
    \hspace{0.02\textwidth} 
    \begin{subfigure}[b]{0.22\textwidth}
        \centering
        \includegraphics[width=\linewidth,height=250pt]{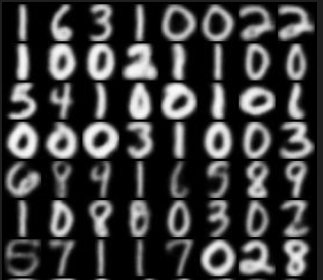} 
        \caption{Q-VAE}
        \label{fig15:quantum_reconstruction_2}
    \end{subfigure}

    \caption{Comparing classical and quantum reconstructed images. (a) USPS input up-sampled image from 16x16 to 32x32 resolution (b) USPS C-VAE image reconstructions, (c) USPS CDP-VAE reconstructed images and (d) USPS Q-VAE reconstructed images.}
    \label{fig:15four_images}
\end{figure}

USPS images reconstructed from 3 different VAE models have been shown in Figure~\ref{fig:15four_images}. For the USPS dataset, while FID score for C-VAE indicate that the model can generate images with reasonable fidelity (i.e., the generated images resemble real USPS digits), it lags behind the performance of the Q-VAE. The quantum model's ability to leverage more complex transformations in the latent space for image reconstruction enables it to capture more intricate patterns and feature representations in the data. This allows the Q-VAE to achieve lower FID scores (indicating higher image fidelity) compared to the CDP-VAE and C-VAE as seen in the results for the USPS dataset. For MSE, the Q-VAE again outperformed the other models, achieving a score of 0.0058. This lower error indicates that the quantum model is particularly effective at capturing the more complex, medium-resolution patterns in the USPS images. The Q-VAE's quantum-enhanced encoding allows it to extract intricate features from the grayscale images, leading to more accurate reconstructions. Its performance on USPS highlights the potential of quantum circuits in improving image generation tasks, especially for datasets with more complex structures compared to MNIST. The C-VAE, with an MSE of 0.0070, performed worse than the CDP-VAE and Q-VAE on USPS, reflecting the same trend observed in MNIST. While still a competitive model, the C-VAE's ability to reconstruct images is limited by its classical encoding mechanism. The model performs well, but its lack of advanced feature extraction techniques prevents it from achieving the same reconstruction quality as the Q-VAE.
The CDP-VAE  with an MSE of 0.0065, performance is better than the C-VAE on MNIST which is 0.0092, it still lags behind both the CDP-VAE and Q-VAE  on USPS. While reconstructed images we can see Q-VAE is reconstructing very well competing with C-VAE and CDP-VAE.

\subsection{New Generated Images}

The performance of the three models in handling random noise varies considerably. The generated images presented were captured at epoch 200. The C-VAE requires more time to converge, necessitating longer training. Therefore, we evaluate the model at epoch 200 to assess its performance after sufficient training. While initial epochs Q-VAE has shown fast convergence at epoch 20, while other models were unable to generate digits even at epoch 20. The Q-VAE outperforms the other models, achieving an FID score of 78.7 at epoch 200. 
In contrast, the C-VAE struggles with random noise, resulting in a much higher FID score of 94.4, indicating poor performance in generating from latent space using Gaussian variable as input. 
The CDP-VAE also faces challenges with noise but performs slightly better than the C-VAE, with an FID score of 93.3. While both the C-VAE and CDP-VAE show limitations in noise handling, the Q-VAE demonstrates a notable advantage in generating more realistic images from random variables. MNIST generated images from noise, have been shown below in Figure \ref{fig16:four_images}:
\begin{figure}[h!]
    \centering
    \begin{subfigure}[b]{0.22\textwidth}
        \centering
        \includegraphics[width=\linewidth,height=250pt]{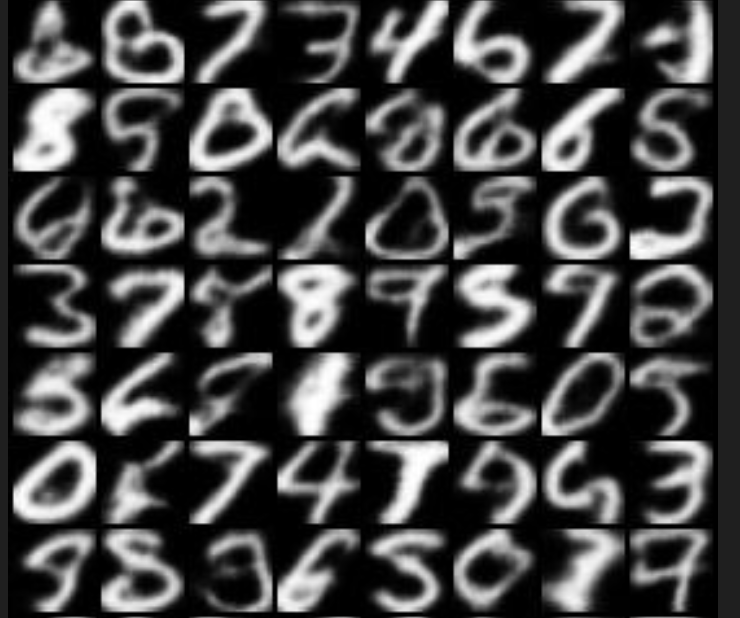}
    \caption{ C-VAE MNIST new generated images}
        \label{fig:classical_reconstruction_1}
    \end{subfigure}    
    \hspace{0.02\textwidth} 
    \begin{subfigure}[b]{0.22\textwidth}
        \centering
        \includegraphics[width=\linewidth,height=250pt]{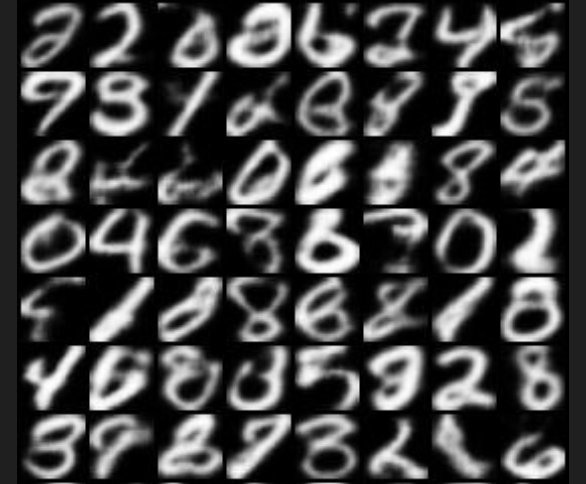} 
        \caption{CDP-VAE new generated images}
        \label{fig:classical_reconstruction_2}
    \end{subfigure}    
    \hspace{0.02\textwidth} 
    \begin{subfigure}[b]{0.22\textwidth}
        \centering
        \includegraphics[width=\linewidth,height=250pt]{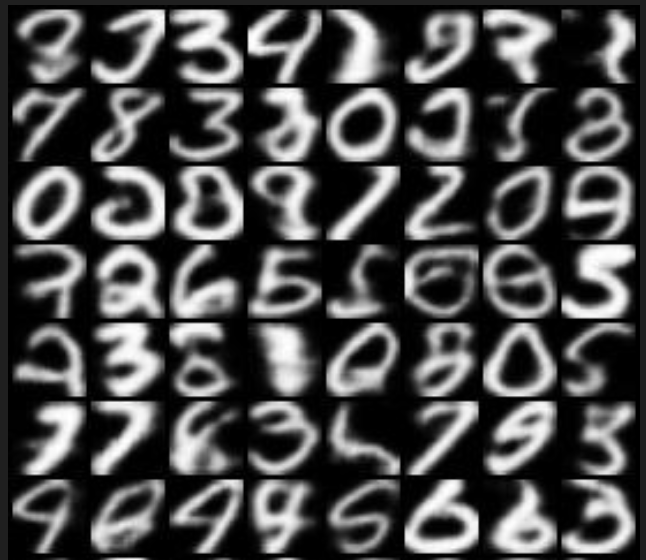} 
        \caption{Q-VAE new generated images}
        \label{fig:quantum_reconstruction_2}
    \end{subfigure}

    \caption{MNIST image generation from Gaussian noise.}
    \label{fig16:four_images}
\end{figure}


These images are organized in a grid format, with varying levels of clarity and detail. While some digits are clearly defined and resemble their expected shapes, others appear blurry, distorted, or lack fine structural details, indicating limitations in the model’s ability to fully capture the underlying data distribution. These artifacts could result from constraints such as an insufficiently expressive latent space, challenges in training stability, or limitations in the model architecture. For instance, models with lower-resolution latent spaces may struggle to encode the nuanced variations required to generate sharp, realistic images.
\begin{figure}[h!]
    \centering
    \begin{subfigure}[b]{0.22\textwidth}
        \centering
        \includegraphics[width=\linewidth,height=250pt]{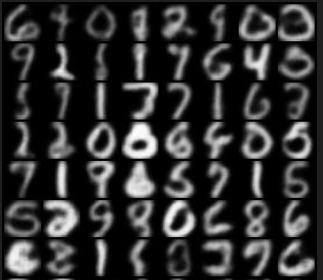}
    \caption{ C-VAE USPS New Generated Images}
        \label{fig:classical_reconstruction_1}
    \end{subfigure}    
    \hspace{0.02\textwidth} 
    \begin{subfigure}[b]{0.22\textwidth}
        \centering
        \includegraphics[width=\linewidth,height=250pt]{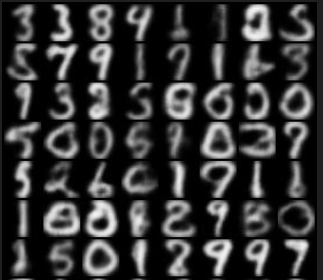} 
        \caption{CDP-VAE New Generated Images}
        \label{fig:classical_reconstruction_2}
    \end{subfigure}    
    \hspace{0.02\textwidth} 
    \begin{subfigure}[b]{0.22\textwidth}
        \centering
        \includegraphics[width=\linewidth,height=250pt]{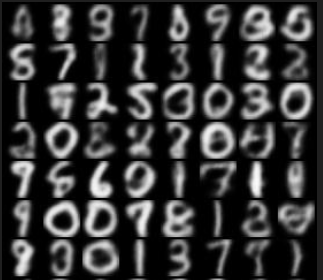} 
        \caption{Q-VAE New Generated Images}
        \label{fig:quantum_reconstruction_2}
    \end{subfigure}

    \caption{USPS image generation from Gaussian noise.}
    \label{fig17:four_images}
\end{figure}
Interestingly, Q-VAE have demonstrated better performance in generating images from noise, as reflected in improved FID scores. 
These operations are highly effective at capturing complex patterns and correlations in medium-resolution data, enabling more detailed and coherent outputs. Similar results were obtained for USPS as shown in Figure~\ref{fig17:four_images}. 



\subsubsection{Trainable Parameters}

The results indicate that both the proposed Q-VAE model and the CDP-VAE demonstrate significant improvements over the C-VAE architecture in terms of parameter efficiency. Specifically, the C-VAE consists of 407,377 trainable parameters, with 299,424 allocated to the encoder and 107,953 to the decoder. In contrast, the CDP-VAE reduces the total number of trainable parameters to 144,977, primarily due to its more compact encoder, which contains only 37,024 parameters. While the Q-VAE also has 144,977 parameters, this reduction is attributed to the CDP-VAE's design rather than the quantum encoding itself. This highlights the role of CDP-VAE in achieving parameter efficiency while leveraging quantum encoding for enhanced feature extraction. Despite the reduction in parameters, the decoder structure remains unchanged across all models, with 107,953 parameters. Notably, the reduction in trainable parameters stems from CDP-VAE, not the quantum component, as the quantum circuit in Q-VAE does not introduce additional trainable parameters. As a result, both CDP-VAE and Q-VAE share the same total number of parameters, totaling 144,977. This demonstrates the efficiency of the CDP approach in reducing parameter overhead while maintaining effective feature extraction.

These findings highlight the role of CDP-VAE in optimizing parameter efficiency within VAE architectures while ensuring high-fidelity image reconstruction. Although C-VAEs remain a widely adopted choice in machine learning, the combination of CDP-VAE’s parameter reduction and Q-VAE’s quantum encoding presents a compelling direction for future research. Potential applications include digital imaging, healthcare diagnostics, and creative arts, where both high-quality image generation and computational efficiency are crucial.
However, the effectiveness and adoption of Q-VAE remain subject to the challenges of quantum hardware scalability and the complexity of quantum algorithms. Continued advancements in quantum machine learning are essential to further explore the integration of quantum encoding in generative models, paving the way for more efficient and powerful solutions in real-world applications.
However, despite these advantages, the adoption and effectiveness of Q-VAE still face challenges, including issues related to the scalability of quantum hardware and the complexity of quantum algorithms. Continued research and development in quantum machine learning are essential for unlocking the full potential of quantum-enhanced models, pushing the boundaries of what is achievable in generative modeling and real-world applications.

\section{Conclusion}

In this study, we introduce the 
quantum down sampling filter for Q-VAE, a novel model that integrates quantum computing into the encoder component of a traditional VAE. Our experimental results demonstrate that Q-VAE offers significant advantages over classical VAE architectures across various performance metrics. In particular, Q-VAE consistently outperforms C-VAE, achieving lower FID scores, indicating superior image fidelity and the ability to generate better-quality reconstructions. The quantum-enhanced encoder captures more intricate features of the images, leading to improved generative performance. We evaluated the model on digit datasets such as MNIST and USPS, initially up-scaled to 32x32 for consistency. For a fair comparison, we also tested our quantum-enhanced model alongside a classical counterpart and a model mimicking classical behavior, referred to as CDP-VAE.
One key finding of this study is the significant reduction in the number of trainable parameters in the CDP-VAE's encoder. While the C-VAE consists of over 407,000 trainable parameters, the CDP-VAE achieves a reduction by significantly lowering the number of parameters. Furthermore, the Q-VAE leverages quantum encoding, which does not add any additional trainable parameters. This combination highlights the efficiency of quantum-based encoding methods and the potential for more scalable and resource-efficient models without increasing computational complexity.
The results highlight the practical advantages of incorporating quantum computing into VAEs, particularly for tasks that require high-quality image generation and feature extraction. The Q-VAE not only improves image quality, but also enhances computational efficiency, making it a promising candidate for future applications in areas such as computer vision, data synthesis, and beyond. These findings suggest that further exploration of quantum-enhanced machine learning techniques could unlock new possibilities in generative modeling and image reconstruction, offering both performance improvements and computational savings.

\vspace{10pt}

\noindent\textbf{Data Availability:}
The data is available at https://github.com/RRFar/Quantum-Downsampling-Filter.

\noindent\textbf{Conflict of Interest:}  
The authors declare no conflict of interest.

\renewcommand{\refname}{References}

\end{document}